\documentclass[11pt]{article}

\usepackage{lipsum}
\usepackage{geometry}
\usepackage{coling2020}
\usepackage{times}
\usepackage{url}

\usepackage{latexsym}
\usepackage{microtype}
\usepackage{graphicx}
\usepackage{graphics}
\usepackage[sorting=none]{biblatex}
\usepackage{array}%需要该宏包

\colingfinalcopy % Uncomment this line for all SemEval submissions

\title{MeisterMorxrc at SemEval-2020 Task 9: Fine-Tune Bert and Multitask Learning for Sentiment Analysis of Code-Mixed Tweets}

\author{Qi Wu \\
  Tianjin University \\ \\ \\
  {\tt Morxrc@163.com} \\ \And
  Peng Wang \\
  University of Electronic  \\ Science  and Technology\\  of China \\
  {\tt wangpeng331439@163.com} \\ \And
  Chenghao Huang \\
  University of Electronic \\ Science  and Technology \\ of China \\
  {\tt zydhjh4593@gmail.com}}
  
\date{}
\newcommand\fn[1]{%
  \begingroup
  \renewcommand\thefootnote{}\footnote{#1}%
  \addtocounter{footnote}{-1}%
  \endgroup
}

\begin{document}

\maketitle
\begin{abstract}
Natural language processing (NLP) has been applied to various fields including text classification and sentiment analysis. In the shared task of sentiment analysis of code-mixed tweets, which is a part of the SemEval-2020 competition~\cite{patwa2020sentimix}, we preprocess datasets by replacing emoji and deleting uncommon characters and so on, and then fine-tune the Bidirectional Encoder Representation from Transformers(BERT) to perform the best. After exhausting top3 submissions, Our team MeisterMorxrc achieves an averaged F1 score of 0.730 in this task, and and our codalab username is MeisterMorxrc.
\end{abstract}

\section{Introduction}
Language is an indispensable and important part of human daily life. Natural language is everywhere as a most direct and simple tool of expression. Natural language processing is to transform the language used for human communication into a machine language that can be understood by machines. It is a model and algorithm framework for studying language capabilities. In recent years, NLP research has increasingly used new deep learning methods. As an important branch of artificial intelligence, language models are models that can estimate the probability distribution of a group of language units (usually word sequences). These models can be built at a lower cost and have significantly improved several NLP tasks, such as machine translation, speech recognition and parsing. The processing flow of natural language can be roughly divided into five steps: obtaining anticipation, preprocessing the corpus, characterizing, model training, and evaluating the effect of modeling.

With the rapid development of the Internet, the frequency of online communication on social software such as Weibo, Twitter, and forums is getting higher and higher, and the Internet itself has also changed from "reading Internet" to "interactive Internet". The Internet has not only become an important source for people to obtain information, but also an important platform for people to express their opinions and share their own experiences and directly express their emotions. The achievements of NLP research laid a good foundation for text sentiment analysis. Text sentiment analysis is an important research branch in the field of natural language understanding, involving theories and methods in the fields of linguistics, psychology, artificial intelligence, etc. It mainly includes the processing of text sources, the subjective and objective classification of network text, and the subjective text Analysis and other steps.

Due to the huge inclusiveness and openness of the Internet itself, it attracts users of different races, different languages, different cultural backgrounds and different religious beliefs to communicate with each other here. Therefore, mixed language sentiment classification will be an important research for NLP direction.

\fn{This work is licensed under a Creative Commons Attribution 4.0 International Licence. Licence details: http://creativecommons.org/licenses/by/4.0/.}

\section{Related Work}
Sentiment analysis is a research with a long history that helps us understand the connections and relationships between objects. In recent years, many scholars have made great progess on sentiment analysis. A basic task in sentiment analysis is classifying the polarity of a given text at the document, sentence, or feature/aspect level—whether the expressed opinion in a document, a sentence or an entity feature/aspect is positive, negative, or neutral. Subsequently, the method described in a patent by Volcani and Fogel~\cite{Volcanietal2001}, looked specifically at sentiment and identified individual words and phrases in text with respect to different emotional scales. Many other subsequent efforts were less sophisticated, using a mere polar view of sentiment, from positive to negative, such as work by Turney~\cite{PeterD.Turneyetal2002}, and Pang~\cite{BoPangetal2002} who applied different methods for detecting the polarity of product reviews and movie reviews respectively. One can also classify a document's polarity on a multi-way scale, which was attempted by Pang~\cite{BoPangetal2005} and Snyder~\cite{Snyderetal2007}.

But according to our findings, this research becomes particularly difficult in multilingual societies, especially in many code-mixed texts. Though some researchers have explored in the field, there is still a long way to go. Sharma and Srinivas explore various methods to normalize the text and judged the polarity of the statement as positive or negative using various sentiment resources~\cite{S.Sharmaetal2015}. Bhargava and Sharma develop a flexible and robust system for mining sentiments from code mixed sentences for English with combination of four other Indian languages (Tamil, Telugu, Hindi and Bengali)~\cite{R.Bhargava2016}. Ghosh and Das extract sentiment (positive or negative) from Facebook posts in the form of code-mixed social media data using a machine learning approach~\cite{SouvickGhoshetal2017}.

\section{Data and Methodology}

\subsection{Data Description}
This task is to predict the sentiment of the mixed tweets of a given code. The sentiment tags are divided into three categories: positive, negative and pertinent. Words are also given unique language tags, such as en (English), spa (Spanish), hi (Hindi), mixed and univ (for example, symbols, @mentioned, hashtags). The given data set is divided into training data set and validation data set, which contains emoticons, symbols, Spanish and English, Hindi and English mixed tweets. Since expressions and symbols cannot be directly put into classification and recognition, preprocessing is required to convert them into recognizable English phrases; for other languages mixed with English, we need to use English phrases to recognize them, and we need to label emotions. So use Multi-task to simultaneously recognize English words and perform emotion prediction.

\subsection{Preprocessing}
We first process the data before feeding the data set to any model. In this section we will introduce the core methods and strategies of processing the raw data.

\textbf{Emoji Substitution} - We design a function to map emoji unicode to replacement phrases. We treat these phrases as regular English phrases so their semantics can be preserved, especially when the dataset size is limited.

\textbf{Character Filtering} - We also convert all the text into lower case. Since 'URL' does not have embedding representation in some pre-trained embedding and models, 'URL' is substituted by 'http'. Besides, we delete some uncommon characters without emotions such as '@' and 'https'.

\subsection{Methodology}

\textbf {Bert} - The input part of BERT is a sequence where two sentences are connected. The two sentences are separated by a separator, and an identification symbol is added to the front and the end of each sentence to indicate the beginning and end of the sentence. For each word, BERT performs three different embedding operations, namely encoding the word position information, Word2vec encoding the word, and encoding the entire sentence. Vector stitching these three embedding results can get BERT input. In order to train the bidirectional feature and obtain the connection between the two sentences, BERT uses the Masked Language Model pre-training method, which randomly covers part of the words in the sentence, and uses the training model to predict this part of the word and the next sentence.This article uses the word vector pre-trained by the BERT method as the vector of the input short text. Because the mixed short texts of tweets are mostly replaced by English words in Spanish and Hindi words or Spanish and Hindi words in English, that is, the context is similar, so Spanish and English, Hindi and English The co-trained word vector is used as the vector for inputting short text.

\textbf{Fine-tune} - The method NFT-TM refers to adding a complex network structure to the upper layer of the BERT model. During training, part of the convolutional layer of the pre-trained model is frozen, the parameters of the BERT are fixed, and only the upper task model network is trained separately. Matthew Peter and others from the Allen Institute for Artificial Intelligence in the United States compared the effects of the FT-TM and NFT-TM methods on the two pre-training models of ELMo and BERT. For BERT, the fine-tune effect is slightly better. Therefore, this experiment uses the NFT-TM method, followed by a simple specific task layer after the Bert model. During training, the BERT is fine-tuned according to the training sample set of the task. The practical results of this task show Excellent results.

\textbf{Multi-task} - Multitask Learning is a derivation transfer learning method. The main tasks use domain-specific information possessed by the training signals of related tasks as a constant task. A machine learning method for inductive bias to improve the generalization performance of main tasks. Multi-task learning involves the simultaneous parallel learning of multiple related tasks and the simultaneous back propagation of gradients. Multiple tasks help each other learn through the underlying shared representation and improve the generalization effect. In this task, due to the mixed sentences of Spanish and English, Hindi and English, and the need to predict emotions.Therefore, the Multi-task model is used to identify English and predict emotion at the same time, reduce network overfitting and improve generalization effect.

\textbf{Adam} - The Adam algorithm is different from the traditional stochastic gradient descent. Stochastic gradient descent keeps a single learning rate to update all the weights. The learning rate does not change during the training process, and Adam calculates the first and second moment estimates of the gradient design independent adaptive learning rates for different parameters. The Adam algorithm records the first moment of the gradient, that is, the average of all gradients in the past and the current gradient, so that each time the update is performed, the gradient of the last update and the current update will not differ much, that is, the gradient is smooth and stable can adapt to unstable objective functions. Adam recorded the second moment of the gradient, that is, the average of the square of the past gradient and the square of the current gradient, which reflects the environmental perception ability, and generates an adaptive learning rate for different parameters. This task uses the Adam algorithm to improve the overall efficiency of solving the problem, and performs more quickly and excellently when the gradient becomes sparse.

\textbf {Tfidf} - We use Tf-Idf to digitize the text in the data set as a string, and then evaluate the importance of a word for a file set or for a corpus where it is located. The importance of a word increases in proportion to the number of times it appears in the document, but at the same time it decreases in inverse proportion to its frequency in the corpus. Both high-frequency words and low file frequencies in the file collection will produce a higher weight TF-IDF, so we use it to filter common words and retain important words, thereby improving the efficiency and accuracy of the overall task.

\section{Experiment Results}
\begin{table}[h]
    \centering
        \begin{tabular}{ccc}
        \hline
            System& F1(averaged)&\\
            \hline
            Fine-tuned Multi-task BERT& 0.730\\
            Fine-tuned BERT& 0.687\\
            Original BERT& 0.589\\
            \hline
        \end{tabular}
    \caption{The results of experiments}
    \label{table:data}
\end{table}
The results on the official test sets for this task are presented in Table 1. Our multi-task BERT model, fine-tuned on the provided dataset.After exhausting top3 submissions, we achieved an averaged F1 score of 0.730.

\section{Conclusion}

In this article, we introduce the results of SemEval-2020 Sentimix Task 9: Sentiment Analysis of Code-Mixed Tweets recognition and classification. The goal of this task is to determine and classify mixed languages, and label mixed sentences of English-Hindi and English-Spanish with positive, negative or neutral emotions. In the task, we first preprocess the data with Emoji Substitution and Character Filtering, convert the emoji recognition into recognizable English words, and then use the Fine-tune method on the processed data to access a specific network after the bert model Layer, through Multi-task and Adam to reduce the network overfitting and improve the generalization effect, improve the efficiency and accuracy of the overall task, and  After exhausting top3 submissions, we achieved a score of 0.730 in this task.

In the future, we will conduct more in-depth research on the entiment Analysis of Code-Mixed Tweets, and use other models and methods to practice and try to continuously improve the stability and accuracy of classification recognition.

% \section{References}
\printbibliography

% include your own bib file like this:
%\bibliographystyle{coling}
%\bibliography{semeval2020}

\end{document}